\documentclass{Interspeech2024}
\usepackage{multirow}
\usepackage{graphicx,xcolor}
\usepackage{CJKutf8}

\interspeechcameraready

\title{Pinyin Regularization in Error Correction for Chinese Speech Recognition with Large Language Models}

\name[affiliation={1}]{Zhiyuan}{Tang}
\name[affiliation={2}]{Dong}{Wang}
\name[affiliation={1}]{Shen}{Huang}
\name[affiliation={1}]{Shidong}{Shang}

\address{
  $^1$Tencent Ethereal Audio Lab, Tencent, China\\
  $^2$Center for Speech and Language Technologies, BNRist, Tsinghua University, China}
\email{atomtang@tencent.com, wangdong99@mails.tsinghua.edu.cn}


\begin{document}

\maketitle

\begin{abstract}
Recent studies have demonstrated the efficacy of large language models (LLMs) in error correction for automatic speech recognition (ASR). However, much of the research focuses on the English language. This paper redirects the attention to Chinese. Firstly, we construct a specialized benchmark dataset aimed at error correction for Chinese ASR with 724K hypotheses-transcription pairs, named the Chinese Hypotheses Paradise dataset (ChineseHP), which contains a wide range of scenarios and presents significant challenges. Subsequently, we conduct a preliminary evaluation using the dataset for both direct-prompting and fine-tuning pre-trained LLMs. Furthermore, we propose a straightforward method of Pinyin regularization for prompts, which involves the transcription of Pinyin directly from text hypotheses. The experimental results reveal that Pinyin regularization consistently enhances the error-correcting ability of LLMs when compared with those without regularization. The dataset is available on the website\footnote{https://github.com/tzyll/ChineseHP}.
\end{abstract}

\section{Introduction}

Automatic speech recognition systems (ASR) are extensively employed in a multitude of applications, including voice search, voice command, and transcription services. Nonetheless, the efficacy of ASR can be significantly influenced by a range of elements, such as background noise, speaker accents, and the fidelity of the audio signal. Errors in the ASR output, particularly in challenging environments, can be adverse to the functionality of downstream applications. Therefore, implementing subsequent error correction processes plays a vital role in enhancing the precision of ASR outputs.

The practice of employing a language model (LM) to rescore the N-best hypotheses from ASR beam search decoding is a common technique to identify the candidate with the lowest perplexity, according to various studies~\cite{mikolov2010recurrent,arisoy2015bidirectional,shin2019effective,yang2021multi,yu2023low}. However, such LM rescoring merely chooses the optimal candidate, thereby neglecting the valuable information contained in the remaining hypotheses. A potentially more advantageous strategy involves merging the N-best hypotheses to generate a new prediction, which is anticipated to be more accurate than the initial candidates~\cite{guo2019spelling,hu2020deliberation,leng2021fastcorrect,hu2022improving,ma2023n,hu2023scaling}.

Recently, large language models (LLMs) have begun to leverage their capacity for understanding language to assist in error correction in a generative style~\cite{chen2023generative,chen2024hyporadise}. In particular, a benchmark for generative error correction (GER) tailored to the English language has been introduced. This benchmark is specifically designed to generate correct transcription directly from N-best hypotheses produced by ASR systems. Additionally, a dataset named HyParadise has been created, containing 316K pairs of hypotheses and transcription, to facilitate the evaluation of GER performance. Building on this, subsequent research has expanded the GER benchmark to encompass a broader range of common noisy conditions encountered in ASR. This expansion has led to the development of a new dataset, Robust HyParadise (RobustHP), comprising 113K pairs for a more comprehensive evaluation of hypotheses and transcription.

While most of existing research primarily focuses on the English language, this study redirects attention to the Chinese language. The objective of this paper is to create a benchmark dataset tailor-made for error correction in Chinese, dubbed the Chinese Hypotheses Paradise dataset (ChineseHP), which comprises 724K hypotheses-transcription pairs. The ChineseHP dataset spans a wide array of contexts and presents a significant challenge for error correction.

Given that Chinese is a logographic language, the pronunciation of its characters does not inherently correspond to their written form. Pinyin, the romanization system for standard Chinese, is extensively utilized throughout China for teaching the language. Additionally, Pinyin is a common method for entering Chinese characters on computers and smartphones. The prevalent use of Pinyin renders it an accessible resource for LLMs to comprehend Chinese.
Moreover, Chinese is rich in homophones, and numerous characters share similar initial or final phonetic elements. Consequently, a Pinyin transcription derived directly from the text hypothesis tends to exhibit a lower error rate than the text hypothesis itself, which is advantageous for the task of error correction.
Therefore, we propose a Pinyin regularization method to be applied both in prompting pre-trained LLMs and during the fine-tuning phase to bolster the stability and efficacy of LLMs in error correction for Chinese ASR. Experimental outcomes consistently demonstrate that Pinyin regularization significantly improves the language model's ability to correct errors in Chinese language materials.

The remainder of the article is structured in the following manner. Section~\ref{sec:chinese-hp} introduces the Chinese Hypotheses Paradise dataset. Section~\ref{sec:pinyin-regularization} details the Pinyin regularization method. Section~\ref{sec:experiments} outlines the experimental framework and findings. The paper is concluded in Section~\ref{sec:conclusion}.

%


\section{Chinese Hypotheses Paradise dataset}
\label{sec:chinese-hp}

The ChineseHP dataset is collected from the ASR outputs of a Chinese-adapted and distilled version of the well-known Whisper Large V2~\cite{radford2023robust}, named Belle-distilwhisper-large-v2-zh\footnote{https://huggingface.co/BELLE-2/Belle-distilwhisper-large-v2-zh}. 

The dataset encompasses four representative Chinese corpora: Aishell-1~\cite{bu2017aishell}, Wenetspeech~\cite{zhang2022wenetspeech}, Aishell-4~\cite{fu2021aishell}, and Kespeech~\cite{tang2021kespeech}. The dataset's statistical details are presented in Table~\ref{tab:dataset-stat}.
It is evident that the dataset contains a diverse range of situations, encompassing regular read speech, broadcast news, meetings, telephone conversations, as well as various accents and dialects.
Specifically, the Aishell-1 corpus comprises standard reading speech, while Wenetspeech contains multiple domains from the internet, and includes the sub-corpora test\_net and test\_meeting for test, which features broadcast news and meeting speech correspondingly. Aishell-4 serves as a telephone conversation corpus, and Kespeech focuses on dialects. Considering Wenetspeech and Kespeech have significantly more data than Aishell-1 and Aishell-4, we limited our sample to 200K utterances from each to maintain balance within the dataset.

For each audio sample, a beam size of 10 was utilized during ASR decoding to generate the top 10 hypotheses. These hypotheses were then converted to simplified Chinese, de-duplicated, and paired with the correct transcriptions to create hypotheses-transcription pairs.

\begin{table}[th]
  \caption{Statistics of the Chinese Hypotheses Paradise dataset (ChineseHP). Wenetspeech/test contains two sub-corpora, i.e., test\_net and test\_meeting.}
  \label{tab:dataset-stat}
  \resizebox{\linewidth}{!}{%
  \centering
  \begin{tabular}{llrrr}
  \toprule
  \multirow{2}{*}{Dataset} & \multicolumn{1}{l}{\multirow{2}{*}{Description}} & \multicolumn{3}{c}{Subset}                                                     \\ \cline{3-5} 
                           & \multicolumn{1}{c}{}                             & \multicolumn{1}{c}{train} & \multicolumn{1}{c}{dev} & \multicolumn{1}{c}{test} \\ \hline
  Aishell-1                & reading style, clean                             & 120,098                   & 14,326                  & 7,176                    \\
  Wenetspeech              & multi-domain, noisy                              & 200,000                   & 13,825                  & 33,143                   \\
  Aishell-4                & meeting, overlapped                              & 97,317                    & 3,959                   & 10,423                   \\
  KeSpeech                 & accent, dialect                                  & 200,000                   & 4,407                   & 19,723                   \\ \hline
  Total                    &                                                  & 617,415                   & 36,517                  & 70,465                   \\
  \bottomrule
  \end{tabular}
  }
  \end{table}

  \begin{table}[th]
    \caption{Character error rate (CER) and Pinyin error rate (PinyinER) of test sets in ChineseHP. Wenetspeech/test contains two sub-corpora, i.e., test\_net and test\_meeting.}
    \label{tab:cer-pinyin}
    \resizebox{\linewidth}{!}{%
    \centering
    \begin{tabular}{lrrrr}
      \toprule
      & \multicolumn{1}{l}{Aishell-1} & \multicolumn{1}{l}{Wenetspeech} & \multicolumn{1}{l}{Aishell-4} & \multicolumn{1}{l}{KeSpeech} \\ \hline
    CER\%      & 5.84                          & 11.97/16.07                     & 25.27                         & 29.83                        \\
    PinyinER\% & 1.46                          & 6.37/11.40                       & 20.37                         & 11.03                       \\
    \bottomrule
    \end{tabular}
    }
    \end{table}

\section{Pinyin regularization}
\label{sec:pinyin-regularization}

\subsection{Pinyin system}
Hanyu Pinyin, often referred to simply as Pinyin, is a romanization system for Mandarin Chinese, typically comprising 23 initials, 24 finals, and 5 tones, including the neutral tone. While various Pinyin systems may exhibit minor variations in the number of initials and finals, they all adhere to the same foundational rules. In this paper, we utilize the Pinyin system as presented in \textit{pypinyin}\footnote{https://pypi.org/project/pypinyin}, with the modification that we employ ``ü" in place of ``v" and ``en" instead of ``n" for the final, as these are more prevalently used in China. The list below displays the initials and finals in Pinyin where the finals include several blends of basic finals, for example, ``ua" is a composite of ``u" and ``a":
{\footnotesize
\begin{itemize}
    \item Initials: b, c, ch, d, f, g, h, j, k, l, m, n, p, q, r, s, sh, t, w, x, y, z, zh
    \item Finals: a, ai, an, ang, ao, e, ei, en, eng, er, i, ia, ian, iang, iao, ie, in, ing, iong, iu, o, ong, ou, u, ua, uai, uan, uang, ue, ui, un, uo, ü, üe
\end{itemize}
}


The pronunciation of Chinese characters is represented through a combination of initials and finals. For instance, the Chinese character for ``you", which is written as ``\begin{CJK*}{UTF8}{gbsn}你\end{CJK*}", is articulated as ``ni3" in Pinyin. Here, ``n" serves as the initial, ``i" acts as the final, and ``3" denotes the tone. It is worth noting that certain Chinese characters can consist solely of a final, without an accompanying initial.
Additional scenarios that may create confusion include:
{\footnotesize
\begin{itemize}
    \item Homophone: different characters have the same pronunciation, some even the same tone, e.g., ``\begin{CJK*}{UTF8}{gbsn}桌\end{CJK*} (desk)" and ``\begin{CJK*}{UTF8}{gbsn}捉\end{CJK*} (catch)" are both pronounced as ``zhuo1".
    \item Heteronym: same character has different pronunciations, e.g., ``\begin{CJK*}{UTF8}{gbsn}都\end{CJK*}" is pronounced as ``dou1 (all)" and ``du1 (capital)".
    \item Alliteration: some initial but different finals, e.g., ``\begin{CJK*}{UTF8}{gbsn}桌\end{CJK*} zhuo1(desk)" and ``\begin{CJK*}{UTF8}{gbsn}助\end{CJK*} zhu4 (help)" have the same initial ``zh".
    \item Rhyme: different initials but same final, e.g., ``\begin{CJK*}{UTF8}{gbsn}都\end{CJK*} dou1 (all)" and ``\begin{CJK*}{UTF8}{gbsn}狗\end{CJK*} gou3 (dog)" have the same final ``ou".
\end{itemize}
}

These various factors can easily perplex ASR systems, leading to mistakes when confronted with any potential disruptions, including noise, accents, or dialects. Regarding errors in ASR output, while the character may be incorrect, the corresponding Pinyin—transcribed directly from the text hypothesis—is frequently accurate, either partially or wholly, often yielding a reduced error rate compared to the text hypothesis alone, as illustrated in Table~\ref{tab:cer-pinyin}. This is clearly advantageous for the task of error correction.


\subsection{Pinyin-regularized prompts}

We have devised a pair of prompt styles: the first is employed for immediate engagement with pre-trained LLMs, such as ChatGPT\footnote{https://chat.openai.com} as referenced in this study, while the second style is tailored for the fine-tuning process of a pre-trained LLM.

\begin{figure}[t]
  \centering
    \includegraphics[trim=130 30 330 80,clip,width=\linewidth]{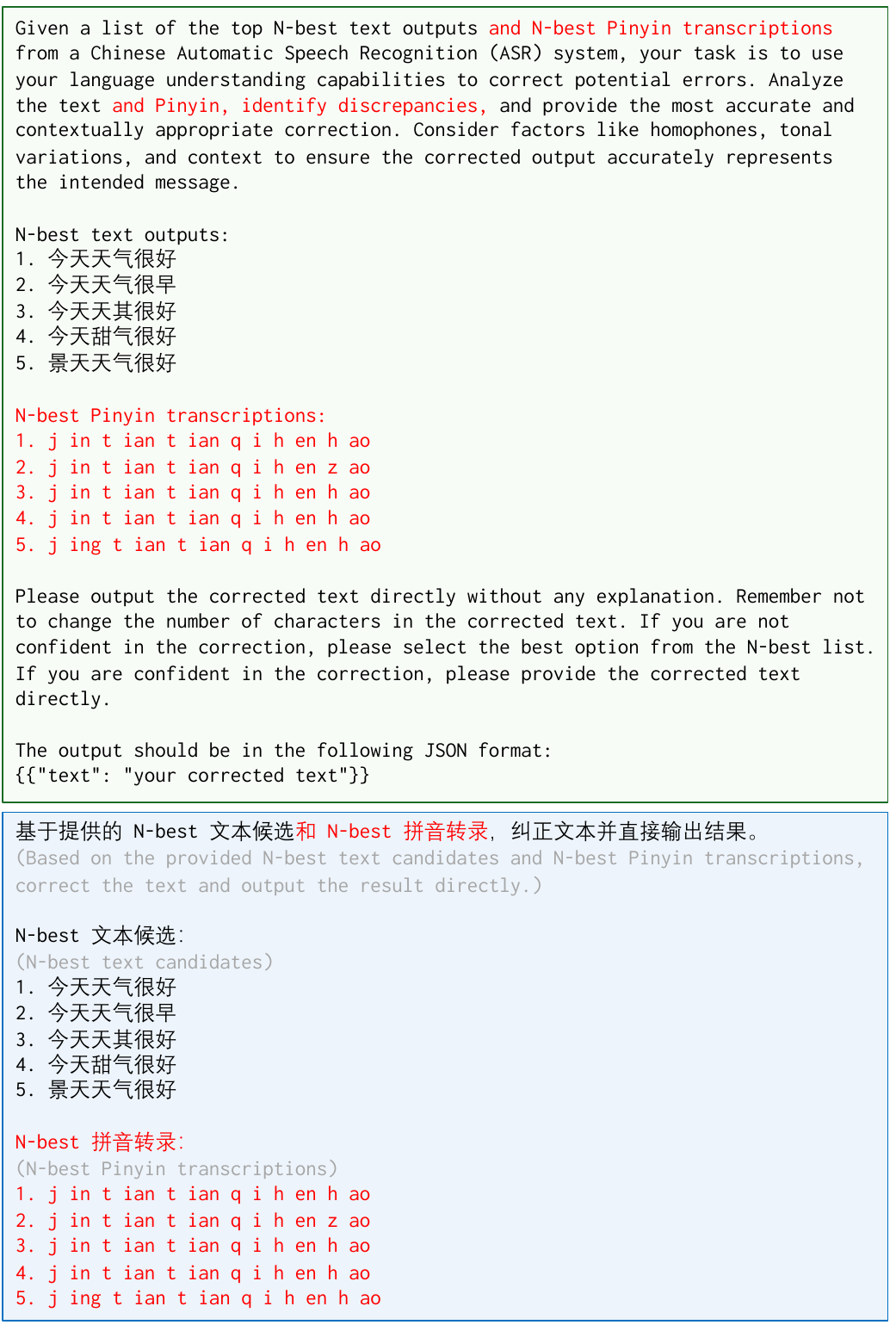}
    \caption{Examples of Pinyin-regularized prompts for direct-prompting ChatGPT (upper) and fine-tuning ChatGLM (lower). The {\color{red}{red}} part is only used when Pinyin regularization is enabled. The {\color{gray}{gray}} part in parentheses is just for translation.
    The corrected text is supposed to be ``\begin{CJK*}{UTF8}{gbsn}今天天气很好\end{CJK*} (Today's weather is very good)".}
    \label{fig:prompt}
  \end{figure}

\subsubsection{Direct prompt}
The first prompt is specifically crafted to engage pre-trained LLMs in a direct manner. In this paper we employ ChatGPT, specifically GPT-3.5. 
Given ChatGPT's proficiency in English, we have utilized an English variant of the prompt as depicted in Figure~\ref{fig:prompt} (upper). 
The prompt is structured to incorporate text hypotheses and the Pinyin transcribed directly from the text hypotheses can be optionally included.
To decrease the hallucination of the language model output, we instruct the model to reply using a JSON format, which gives more stability and controllability.

\subsubsection{Prompt in fine-tuning}
As for fine-tuning LLMs for Chinese error correction, this paper favors the chatGLM model~\cite{du2022glm}, given its considerable focus on the Chinese language. The training data is formatted in a prompt-response manner, which involves incorporating the pairs of hypotheses and transcription from ChineseHP into the Chinese prompts, as illustrated in Figure~\ref{fig:prompt} (lower).


\section{Experiments}
\label{sec:experiments}
\subsection{Direct-prompting ChatGPT}
We perform experiments to assess how various prompts affect the performance of ChatGPT in error correction. 
Taking into account the number of optimal hypotheses for either text or Pinyin, along with the kinds of Pinyin, we have crafted 9 distinct prompts. These are detailed in Table~\ref{tab:prompt-gpt}.

\begin{table}[th]
  \caption{Different prompts for ChatGPT. ``${*}$" means repeating the first candidate for N times. Ground-truth Pinyin is only used for analysis and unavailable in practice.}
  \label{tab:prompt-gpt}
  \centering
  \resizebox{0.7\linewidth}{!}{%
  \begin{tabular}{lccc}
  \toprule
  \multicolumn{1}{c}{\multirow{3}{*}{\textbf{Method}}} & \multicolumn{3}{c}{\textbf{N for best hypotheses}}         \\ \cline{2-4} 
  \multicolumn{1}{c}{}                                 & \multirow{2}{*}{Text} & \multicolumn{2}{c}{Pinyin}         \\ \cline{3-4} 
  \multicolumn{1}{c}{}                                 &                       & Transcribed         & Ground-truth \\ \hline
  Baseline                                             & -                     & -                   & -            \\ \hline
  Prompt1                                              & 5                     & -                   & -            \\
  Prompt2                                              & 5                     & 5                   & -            \\
  Prompt3                                              & 5                     & -                   & 1            \\
  Prompt4                                              & 5                     & -                   & \ \ 5${^*}$  \\ \hline
  Prompt5                                              & 1                     & -                   & -            \\
  Prompt6                                              & 1                     & 1                   & -            \\
  Prompt7                                              & 1                     & -                   & 1            \\ \hline
  Prompt8                                              & \ \ 2${^*}$    & -                   & -            \\
  Prompt9                                              & 1                     & \ \ 2${^*}$  & -            \\
  \bottomrule
  \end{tabular}
  }
  \end{table}

  \begin{figure}[h]
    \centering
      \includegraphics[trim=15 10 70 60,clip,width=\linewidth]{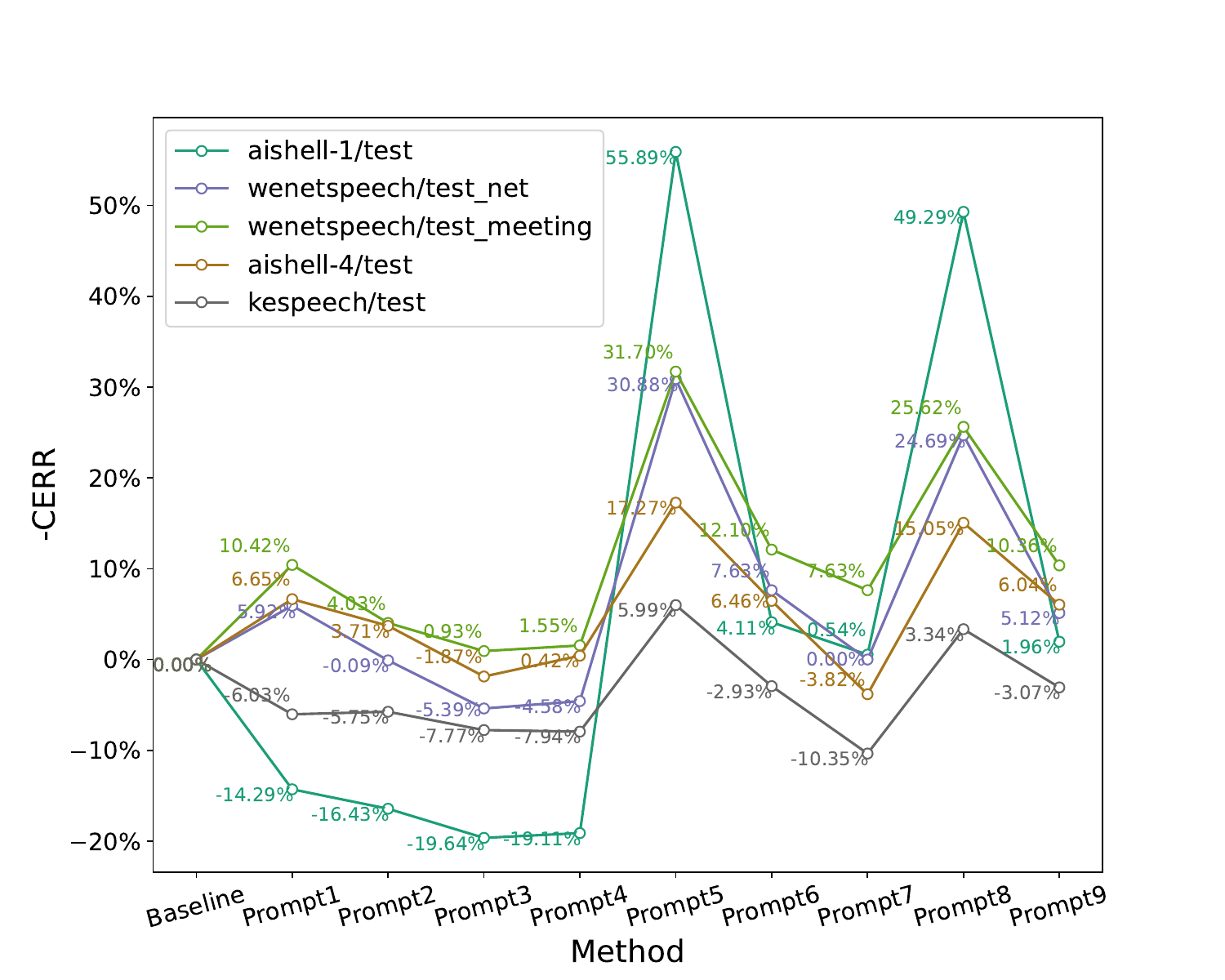}
      \caption{Minus Character error rate reduction (-CERR) of different prompts for ChatGPT. Lower is better.}
      \label{fig:cerr-prompt-gpt}
    \end{figure}

Considering economy and efficiency, we selectively sampled approximately 2,000 hypotheses-transcription pairs from the test set of each ChineseHP dataset to assess the efficacy of various prompts. We use the character error rate reduction (CERR) to measure the improvements different methods offered over the baseline. The variations in CERR across different prompts for each dataset are depicted in Figure~\ref{fig:cerr-prompt-gpt}.

The figure illustrates that incorporating Pinyin within the prompt enhances the performance of ChatGPT, with accuracy directly correlating to the precision of the provided Pinyin, as indicated by $prompt3>prompt2>prompt1$ ($>$ means `better than', same below).
This effect is particularly notable when considering a singular best hypothesis, as demonstrated by $prompt7>prompt6>prompt5$.

In an attempt to discern whether duplicating the 1-best text hypothesis could serve as an effective alternative to Pinyin, we carry out an experiment, but the results indicated that this method was less effective when compared to the use of Pinyin, as shown by $prompt6>prompt8$.

Moreover, we double the 1-best Pinyin hypothesis to determine the effect of additional Pinyin in the prompt. This leads to a modest enhancement in performance, with $prompt9>prompt6$, although the gain was not as substantial as the initial inclusion of Pinyin.

Inspired by these findings, we experiment with replicating the ground-truth Pinyin hypothesis 5 times. However, this results in only a negligible improvement in performance: $prompt4 \approx prompt3$. This suggests that the 5-best text hypotheses, providing a variety of information, may offer a comparable regularization effect and that further repetition of the same Pinyin does not contribute significant additional insight.

\begin{table*}[ht]
  \caption{Results of different fine-tuning prompts for error correction with ChatGLM. 
  $o_{cp}$ and $o_{nb}$ are compositional oracle and n-best oracle respectively, defined in Section~\ref{sec:finetune}.
  The base model is unable to perform the task precisely, so the CERs are not available.}
  \label{tab:res-finetune}
  \centering
  \resizebox{\textwidth}{!}{%
  \begin{tabular}{lccclllll}
  \toprule
  \multirow{2}{*}{\textbf{Method}} & \multicolumn{2}{c}{\textbf{N for best hypotheses}} &  & \multicolumn{5}{c}{\textbf{CER\%}$\downarrow$ $_{-CERR\%\downarrow}$}                                                                                                                                                                \\ \cline{2-3} \cline{5-9} 
                                   & Text                     & Transcribed Pinyin      &  & \multicolumn{1}{l}{Aishell-1/test} & \multicolumn{1}{l}{Wenetspeech/test\_net} & \multicolumn{1}{l}{Wenetspeech/test\_meeting} & \multicolumn{1}{l}{Aishell-4/test} & \multicolumn{1}{l}{KeSpeech/test} \\ \hline
Baseline                         & -                        & -                       &  & 5.84                                             & 11.62                                       & 16.03                                         & 25.19                                                      & 29.81                                                     \\
$o_{cp}$                         & -                        & -                       &  & 2.93{\color{teal} $_{-49.83}$ }                & 6.89{\color{teal} $_{-40.71}$ }           & 8.67{\color{teal} $_{-45.91}$ }             & 13.34{\color{teal} $_{-47.04}$ }                         & 22.9{\color{teal} $_{-23.18}$ }                         \\
$o_{nb}$                         & -                        & -                       &  & 3.49{\color{teal} $_{-40.24}$ }                & 8.48{\color{teal} $_{-27.02}$ }           & 12.76{\color{teal} $_{-20.40}$ }            & 20.06{\color{teal} $_{-20.37}$ }                         & 25.49{\color{teal} $_{-14.49}$ }                        \\ \hline
Finetune1                        & 5                        & -                       &  & 5.14{\color{teal} $_{-11.99}$ }                & 13.27{\color{gray} $_{14.20}$ }          & 17.3{\color{gray} $_{7.92}$ }              & 27.4{\color{gray} $_{8.77}$ }                           & 31.12{\color{gray} $_{4.39}$ }                         \\
Finetune2                        & 5                        & 5                       &  & 4.74{\color{teal} $_{-18.84}$ }                & 12.28{\color{gray} $_{5.68}$ }           & 16.33{\color{gray} $_{1.87}$ }             & 25.61{\color{gray} $_{1.67}$ }                          & 28.73{\color{teal} $_{-3.62}$ }                         \\
Finetune3                        & 1                        & -                       &  & 6.88{\color{gray} $_{17.81}$ }                & 14.36{\color{gray} $_{23.58}$ }          & 18.25{\color{gray} $_{13.85}$ }            & 28.09{\color{gray} $_{11.51}$ }                         & 33.25{\color{gray} $_{11.54}$ }                        \\
Finetune4                        & 1                        & 1                       &  & 6.26{\color{gray} $_{7.19}$ }                 & 14.06{\color{gray} $_{21.00}$ }          & 17.94{\color{gray} $_{11.92}$ }            & 27.61{\color{gray} $_{9.61}$ }                          & 31.2{\color{gray} $_{4.66}$ }                          \\
  \bottomrule
  \end{tabular}
  }
  \end{table*}

\subsection{Fine-tuning ChatGLM}
\label{sec:finetune}

We perform a series of experiments to assess the efficacy of various prompts for fine-tuning ChatGLM (Version 3\footnote{https://github.com/THUDM/ChatGLM3} with 6 billion parameters).
Taking into account the optimal number of hypotheses for text or Pinyin, we have formulated four distinct fine-tuning prompts, which are detailed in Table~\ref{tab:res-finetune}.

The training data is prepared by combining subsets of each dataset in ChineseHP, precisely 10K hypotheses-transcription pairs from Aishell-1/train, 20K from Wenetspeech/train, 20K from Aishell-4/train, and 20K from KeSpeech/train. The data list can be found on the ChineseHP website.
We use 4 NVIDIA A100 GPUs to fine-tune the ChatGLM model with 2 epochs for each fine-tuning prompt.
For computational efficiency, the low-rank adaptation (LoRA) approach~\cite{hu2021lora} is applied to optimize a small number of parameters.
The fine-tuning pipeline and LoRA configuration follows an open-source repository\footnote{https://github.com/liucongg/ChatGLM-Finetuning}.

The results are presented in Table~\ref{tab:res-finetune}.
Due to the limited number of instances in training data with transcriptions exceeding 100 characters, we confine our report to test set results with lengths under 100 characters concerning references, hypotheses, and LLM outputs to prevent hallucination.
As in~\cite{chen2024hyporadise}, we also provide a comparison with two oracle CERs for reference, namely: 1) the n-best oracle $o_{nb}$, which represents the CER of the  ``best candidate" in the N-best hypotheses, and 2) the compositional oracle $o_{cp}$, indicating the minimum achievable CER by utilizing ``all tokens" available in the N-best hypotheses. The $o_{nb}$ is indicative of the potential peak performance of re-ranking-based approaches, whereas $o_{cp}$ signals the ceiling for corrections that utilize elements present in the list.

The table reveals that fine-tuning ChatGLM with 5-best text hypotheses leads to a marked performance enhancement on the Aishell-1/test dataset. However, the improvement in performance on more intricate settings, such as Wenetspeech/test, Aishell-4/test, and KeSpeech/test, remains minimal. This aligns with the findings from ChatGPT, highlighting the significant challenges associated with error correction in these scenarios.
The promising aspect is that the base model is empowered to perform the task precisely via fine-tuning.

The performance observed with the 1-best text or Pinyin hypothesis is not as good as that achieved with ChatGPT. We conjecture that this is due to the inadequate model size and the limited amount of training data, which fail to capture the complementary information, leading to a tendency for the model to overfit to the training dataset.

On the other hand, ChatGLM's performance is consistently enhanced when Pinyin is incorporated into the fine-tuning process, regardless of whether the number of top hypotheses is 5 or 1. This is evidenced by the fact that $finetune2 > finetune1$ and $finetune4 > finetune3$. Such results indicate the promising role of Pinyin regularization in reducing the errors in Chinese ASR systems.

\subsection{Case Analysis}

\begin{CJK*}{UTF8}{gbsn}
\begin{table}[]
  \caption{Case analysis of different fine-tuning prompts for error correction with ChatGLM. The {\color{red}red} Pinyin is for the errors compared to the ground truth.}
  \label{tab:case}
  \centering

  \resizebox{\linewidth}{!}{%
  \begin{tabular}{llr}
  \toprule
  \textbf{Method}              & \textbf{Utterance}                                     & \multicolumn{1}{l}{\textbf{CER\%$\downarrow$}} \\ \hline
                               & \multicolumn{1}{c}{Case 1}                                            &               \\
  Ground truth                 & 一线楼市成交量激{\color{red}ji1}增{\color{red}zeng1}                                        & -              \\  \hline
  \multirow{5}{*}{N-best list} & 一线楼市成交量基{\color{red}ji1}增                                               & 11.11                              \\
                               & 一线楼市成交量机{\color{red}ji1}增                                               & 11.11                              \\
                               & 一线楼市成交量积{\color{red}ji2}增                                               & 11.11                              \\
                               & 一线楼市成交量基{\color{red}ji1}僧{\color{red}seng1}                                        & 22.22                              \\
                               & 一线楼市成交量基{\color{red}ji1}升{\color{red}sheng1}                                       & 22.22                              \\ \hline
  Finetune1                    & 一线楼市成交量继续保持                                                & 44.44                              \\
  Finetune2                    & 一线楼市成交量激增                                                    & 0.00              \\
  Finetune3                    & 一线楼市成交量激增                                                    & 0.00              \\
  Finetune4                    & 一线楼市成交量即增                                                    & 11.11                              \\ \hline \hline
                               & \multicolumn{1}{c}{Case 2}                                            &               \\
  Ground truth                 & {\scriptsize 当{\color{red}dang1}你{\color{red}ni3}面对马凝{\color{red}ning2}视{\color{red}shi4}时{\color{red}shi2}马则不愿前{\color{red}qian2}行    }& -              \\ \hline
  \multirow{5}{*}{N-best list} & {\scriptsize 但{\color{red}dan4}念{\color{red}nian4}面对马宁{\color{red}ning2}事{\color{red}shi4}实{\color{red}shi2}马则不愿意{\color{red}yi4}牵{\color{red}qian1}行 }& 50.00                                 \\
                               & {\scriptsize 但{\color{red}dan4}你面对马宁{\color{red}ning2}事{\color{red}shi4}实{\color{red}shi2}马则不愿意{\color{red}yi4}牵{\color{red}qian1}行        }& 42.86                              \\
                               & {\scriptsize 当年{\color{red}nian2}面对马宁{\color{red}ning2}事{\color{red}shi4}实{\color{red}shi2}马则不愿意{\color{red}yi4}牵{\color{red}qian1}行              }& 42.86                              \\
                               & {\scriptsize 但{\color{red}dan4}念{\color{red}nian4}面对马宁{\color{red}ning2}事{\color{red}shi4}实{\color{red}shi2}马则不愿意{\color{red}yi4}谦{\color{red}qian1}行 }& 50.00                                 \\
                               & {\scriptsize 但{\color{red}dan4}你面对马宁{\color{red}ning2}事{\color{red}shi4}实{\color{red}shi2}马则不愿意{\color{red}yi4}谦{\color{red}qian1}行        }& 42.86                              \\ \hline
  Finetune1                    & 但面对马宁实事求是地回答了问题                                        & 92.86                              \\
  Finetune2                    & 但面对马宁试马则不愿意前进一步                                        & 64.29                              \\
  Finetune3                    & 但面对马赛时马则不愿意前进一步                                        & 57.14                              \\
  Finetune4                    & 但面对马宁失事马则不愿意签新                                          & 57.14                             \\
  \bottomrule
  \end{tabular}
  }
  \end{table}
  \end{CJK*}

Two cases are shown in Table~\ref{tab:case} to illustrate the performance of different fine-tuning prompts for error correction with ChatGLM. Case 1 is from Aishell-1/test, which is a standard reading sample with few errors in the N-best list, while Case 2 is from KeSpeech/test, which has more errors due to the accent.

In Case 1, the one or two errors can be effectively corrected with Pingyin regularization even with the 1-best hypothesis, while the result of the 5-best text hypotheses without Pinyin gets some hallucination.

In Case 2, the errors are of a more complicated nature, making it challenging to recover the original intent from the given hypotheses. This complexity increases the probability of hallucination, leading to a decline in performance across all fine-tuning models. Nonetheless, Pinyin regularization continues to mitigate the issue, preventing the partial text from being incorrectly re-generated under the constraints of the Pinyin.

\section{Conclusion}
\label{sec:conclusion}

In this study, we introduce a specialized benchmark dataset designed for error correction in Chinese ASR named the Chinese Hypotheses Paradise dataset (ChineseHP), comprising 724K hypotheses-transcription pairs. 
The dataset covers a wide range of real-world scenarios, representing a considerable challenge. 
Additionally, we develop baseline methods for prompting and fine-tuning pre-trained LLMs and propose a simple yet effective Pinyin regularization technique to enhance their robustness and performance. 
Moving forward, we plan to investigate more sophisticated fine-tuning approaches, develop more potent prompts, and utilize additional training data to further enhance the error correction capabilities of LLMs for Chinese ASR.

%

\bibliographystyle{IEEEtran}
\bibliography{mybib}

\end{document}